# Context-Aware Single-Shot Detector

Wei Xiang[2]   Dong-Qing Zhang[1]   Heather Yu[1]   Vassilis Athitsos[2]
[1]Media Lab, Futurewei Technologies   [2]University of Texas at Arlington
wei.xiang@mavs.uta.edu, {dongqing.zhang, heatheryu}@huawei.com, athitsos@uta.edu

## Abstract

*SSD (Single Shot Detector) is one of the state-of-the-art object detection algorithms, and it combines high detection accuracy with real-time speed. However, it is widely recognized that SSD is less accurate in detecting small objects compared to large objects, because it ignores the context from outside the proposal boxes. In this paper, we present CSSD–a shorthand for context-aware single-shot multibox object detector. CSSD is built on top of SSD, with additional layers modeling multi-scale contexts. We describe two variants of CSSD, which differ in their context layers, using dilated convolution layers (DiCSSD) and deconvolution layers (DeCSSD) respectively. The experimental results show that the multi-scale context modeling significantly improves the detection accuracy. In addition, we study the relationship between effective receptive fields (ERFs) and the theoretical receptive fields (TRFs), particularly on a VGGNet. The empirical results further strengthen our conclusion that SSD coupled with context layers achieves better detection results especially for small objects ($+3.2\%\text{AP}_{@0.5}$ on MS-COCO compared to the newest SSD), while maintaining comparable runtime performance.*

## 1. Introduction

Deep learning approaches have shown some impressive results in general object detection. However, there still remain fundamental challenges to be addressed, particularly in detecting objects of dramatically different scales. In the literature, many attempts have been made to overcome this issue: from *image pyramids*-based approaches which have often been combined with hand-crafted features [19, 3] to *feature map pyramid*-based approaches [25, 17] within a deep learning framework. In addition, the state of the art has moved from the *sliding-window* paradigm to the much more efficient alternative of *feature maps scanning*, thanks to the representation and learning capabilities of Convolutional Neural Networks (CNNs).

Within the deep learning paradigm, methods predicting proposals from feature maps of the single highest-level scale [8, 10, 25] enable the most variance of scales, due to the shared semantics. However, such methods also suffer from slow inference time, given that the hypothesized proposals at all scales need to go through the whole CNN, leading to large computational cost and memory usage. Single Shot Detector (SSD) [17] mitigated this issue by making predictions from feature maps of multiple scales in a hierarchical approach. This way, when hypothesizing proposals at increasing scales, SSD goes deeper in CNNs with more learnable parameters and thus takes longer time. Due to this *bottom-up* design, SSD assumes that small object detection only relies on fine-grained local features, and ignores context outside those local features. In Fig. 1, we show, for an SSD detector, the only *conv4_3* Receptive Field (RF) that was used to detect the smallest object. Within that RF, we can hardly recognize the object of interest (green box). After expanding the *conv4_3* RF into multiple higher scales, more visual cues (marked by red ellipses) become available. With that contextual information, it is possible not only to recognize the object as a sheep, but also to detect the presence of a herd in the given picture.

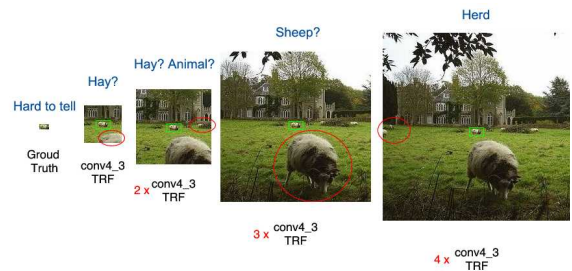

Figure 1: In SSD [17], *conv4_3* of the VGG16 was used to detect the smallest object. As the TRF of *conv4_3* includes limited visual cues, integrating informative context from different scales can help us detect very small objects.

Both two dominant deep learning object detection methods: Faster R-CNN [25] and SSD [17], require precomputed grids (named anchors in [25] and default boxes in [17])[1] to either generate proposals or regress and clas-sify directly upon them. Most of the previous work in deep learning object detection focuses on architecture design of networks, but there is relatively little work studying the underlying instrument of generating grids and proposals. Consequently, the hierarchical structures in networks and grid scales have to be fine-tuned exhaustively in order to obtain satisfactory results.

---
[1]We use grid and default box interchangeably throughout this paper.

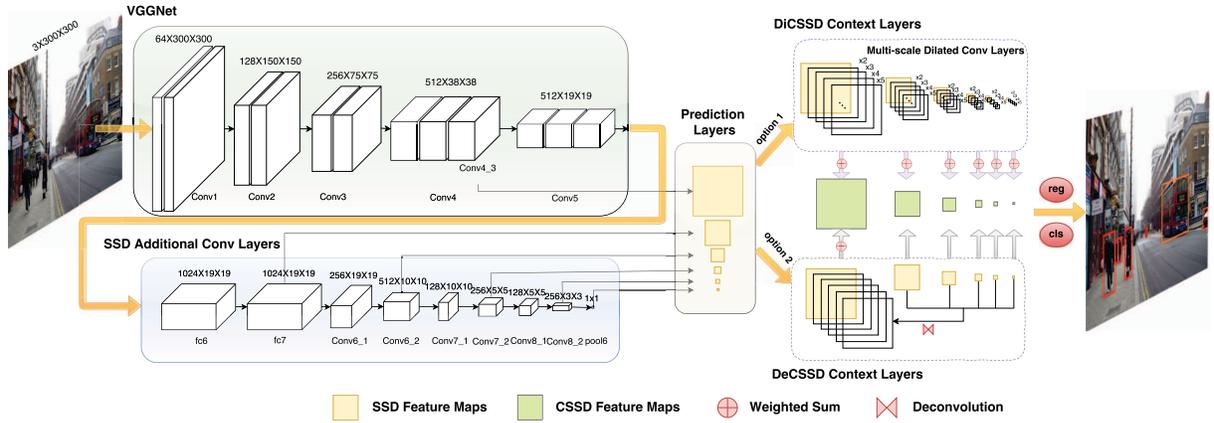

Figure 2: Flow diagram of CSSD. We built CSSD directly upon SSD with two implementations of context layers using 1) multi-scale dilated convolution layers, 2) deconvolution layers. The former has an adjustable parameter controlling the number of context layers, while the latter directly reuses the feature maps from upper layers fused specifically for smallest object detection. Both implementations learn associated scaling parameters during fusion of feature maps.

In the VGG16 version[2] of SSD, prescribed grid scales associated with each RF size were specifically designed at different prediction layers, to ensure that every point on the feature maps of prediction layers *sees* a sufficiently large area from input image. Take as an example a $300 \times 300$ input: the grid scales versus RF sizes at all SSD's 6 prediction layers are 30/92, 60/420, 114/452, 168/516, 222/644 and 276/644 respectively. However, according to [33, 20], the effective receptive fields (ERFs) are 2D-gaussian distributed and proved to be significantly smaller than the corresponding theoretical receptive fields (TRFs).

Following [33], our analysis (Fig. 5(c)) shows that the ERF of *conv4_3* in SSD is only $\times 1.9$ larger than the corresponding grid scale $\theta_p$ (not exceeding $\times 2.5$ across all prediction layers), which motivates the need for more contexts to be integrated into the existing framework. In this paper, we present a context-aware framework for SSD (Fig. 2), and give two different implementations, which either expands RF sizes with multiple scales and then fuses to form new prediction layers, or reuses directly feature maps from higher layers to integrate complex structure features. Because the scale parameters are automatically learned during fusion of contextual feature maps, the proposed network enables the trade-off between fine-grained features and richer features encompassing more context.

In summary, this paper makes the following main contributions:

(1) To our best knowledge, we are the first to analyze ERFs within the framework of object detection. Using our analysis, we provide the ERF sizes of a standard VGG16 network [27]. These sizes can be utilized as a reference to design more effective CNNs for example in object detection.

(2) We present a new framework based on SSD, by introducing multi-scale dilated convolution layers in a hierarchical approach, named Dilation-based Context-aware SSD (DiCSSD). Moreover, we alternatively propose a VGG16-based, deconvolutional version of context-aware SSD (DeCSSD). Both were designed specifically for small-scale object detection with scaling parameters learned automatically during feature map fusions. Our experimental results on VOC [5], MS-COCO [16] and DETRAC [30] show that both DiCSSD and DeCSSD outperform SSD while maintaining real-time speed, in addition to producing promising detection results on small objects. The evaluation code and trained models are publicly available.[3]

## 2. Related Work

**Early object detectors in CNNs.** Until a few years ago, the sliding-window paradigm [4, 29] was commonly used. With the emerging technology of deep learning, G. Ross et al. [9] proposed an object detection framework based on deep neural networks named R-CNN, which performs a forward pass for every object proposal and then classifies. Its followers, SPPnet [10] and Fast R-CNN [8] have been shown to largely reduce the training and inference time by sharing computation upon convolutional feature maps for the entire input image.

---

[2]We compare the performance of ResNet50/101/152 versions of SSD as well in our experiments. Considering that the runtime performance of SSD is dramatically reduced with ResNet101/152, in this paper we mainly investigate context layers for the VGG16 version of SSD.

[3]https://github.com/xw1120/CSSD

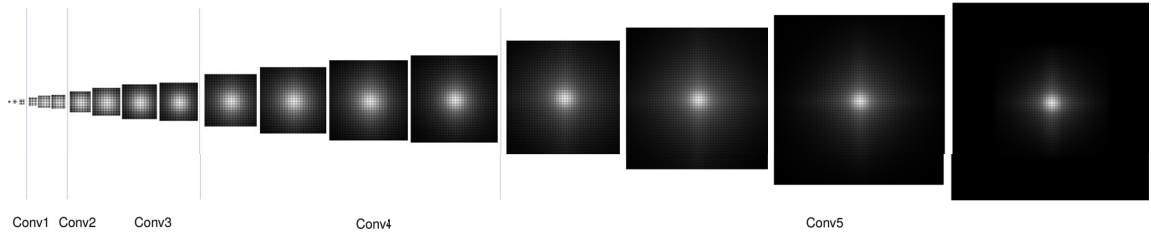

Figure 3: Visualization of ERFs vs. TRFs in a VGG16 network.

**Faster R-CNN.** With the introduction of Region Proposal Network (RPN), the seminal work Faster R-CNN [25] unblocked the bottleneck imposed by hypothesizing region proposals that took a large amount of processing time in Fast R-CNN. Despite their crucial dependency on the proposal generation methods/networks, variants of Fast(er) R-CNN[4][32, 14] advanced the performance to a great extent in a majority of open datasets like vehicle detection datasets KITTI [7], DETRAC [30], and some general object detection datasets like PASCAL VOC [5] and MS-COCO [16].

More recent methods based on Faster R-CNN framework include Feature Pyramid Network (FPN) [15] which designs a generic architecture with lateral connections between low-resolution feature maps and higher, and Region-based Fully Convolutional Networks (R-FCN) [14] which focuses on translation-invariance detections.

**Single-shot detectors.** Most Fast(er) R-CNN methods are able to classify ROIs in 200ms, however, proposal generation takes much longer than that and becomes a major bottleneck in deploying the trained model into real-time systems. Therefore, by directly sampling grids upon the input image, and then training a CNN to directly regress and classify (namely single-shot prediction), method You Look Only Once (YOLO) [23] accelerated the runtime performance to 45fps. However, YOLO incurred a non-negligible penalty in accuracy, due to the coarseness of the features that were learned. One of the state-of-the-art single-shot prediction methods, SSD [17], alleviated this problem with an extension of VGG16 network used to predict multi-scale grids (default boxes named in [17]) in a hierarchical structure. Despite having distinct advantages over Fast(er) R-CNN, SSD suffers in accuracy at the task of detecting small objects, because it cannot integrate context for the features learned in bottom layers. We note that context has been shown crucial in recognizing tiny objects [13]. Furthermore, as the input size of SSD is limited below $\sim$512 (otherwise its number of default boxes grows exponentially, incurring overwhelming computational and memory cost), its performance on small object detection is greatly constrained.

More recently, YOLO9000 [24] improves accuracy by integrating dataset-specific data augmentation techniques, which brings the need for more hyper-parameters. DSSD [6] is also built directly on the existing SSD framework, and therefore is the most similar to our proposed method. However, DSSD opts for ResNet101 as its base network, and therefore its runtime performance has dropped from 46fps to 10fps. Both the proposed CSSD and DSSD were designed to improve small object detection. The proposed CSSD method achieves real-time speed and has been shown to perform significantly better than SSD, in both MS-COCO [16] and vehicle detection dataset DETRAC [30].

**Effective Receptive Fields.** Receptive field (RF), also called *field of view*, refers to a region that a unit neuron in a certain layer of the network sees/depends on in the original input image. Most previous work utilized the theoretical receptive field sizes to guide their network designs [28, 21]. However, those designs have not considered whether each pixel in the TRF contributes equally to the output of a certain unit neuron. Zhou et al. [33] were the first to introduce the concept of *empirical receptive fields*, and showed via a data-driven approach that the actual size of RF is much smaller than TRF. Nevertheless, a solid mathematical model of how the empirical receptive fields relate to their theoretical counterparts was offered only recently, by Luo et al. [20], who pointed out that not all pixels in the *effective receptive fields*[5] contributes equally to a unit neuron's response. Instead, the centers of RFs have much larger impacts on the output leading to an obvious 2D-gaussian shape. This is because in the forward pass the central pixels can reach the output with many more different paths than the pixels in outer area, while in the backward pass gradients are back-propagated across all paths equally [20].

---

[4]In this paper, we refer to *Fast(er) R-CNN* as the CNNs that require region proposals using either RPN or external proposal methods.

[5]Empirical receptive fields and effective receptive fields are used interchangeably in this paper.

Inspired by [33], we calculate the ERF sizes across all VGG16 layers in SSD with associated TRF sizes (Fig. 3). To our best knowledge, we are the first to provide the ERF sizes of a standard VGG16 Network [27]. In addition, we compare our computed results against the fitted ERF sizes (Table 1) with $\sqrt{\text{TRF}}$, whose finding suggests us to introduce the proposed context layers, which are fused so as to enable the trade-off between fine-grained features and richer features encompassing more context.

## 3. Context-Aware Single-Shot Detector

We show the framework of our proposed CSSD in Fig. 2. Both CSSD and SSD[6] utilize a standard VGG16 [27] as its base network. However, CSSD uses the feature maps of prediction layers after fusion of context layers, rather than using them directly as in SSD.

To integrate context for small object detection, intuitively we reuse the feature maps from top layers and merge them into the bottom one. However, as the feature map sizes differ a lot (for e.g. by a factor of 1/2 after every pooling layer with $\text{stride} = 2$), we need to *upsample* maps from top layers first and ensure all sizes are the same. We can consider the convolutional part of the network as an *encoding* step, and deconvolution layers can be treated as a *decoding* processing step. The encode-decode structure is known as an *hourglass* and has been shown to be particularly useful in segmentation [22]. As SSD branches out into different prediction layers, in our design of DeCSSD (option 2 in Fig. 2), we make hourglasses from all prediction layers except the first one (where the hourglass is essentially the layer itself) and fuse them with learnable scaling parameters across all maps. Subsequently, the fused feature map becomes the new prediction layer used for small object detection only.

Using deconvolution layers to integrate multi-scale contexts has one obvious drawback, namely that the memory usage of network increases significantly, because the coefficients of bilinear filters and the following conv layers (which have been verified to be especially useful in our experiments) take many more weight parameters. Besides, DeSSD does not always work *out of the box*. Due to that, the training error during fusion of *conv4_3's* context layers is prone to drifting. To address that issue, we additionally add one batch normalization layer that allows for more stable learning after each context layer.

---

[6]In the newest implementation of SSD [18], its additional conv layers have been extended further with more parameters, but the number of prediction layers is unchanged. Also note that any similar extensions to SSD can easily be incorporated into CSSD.

We alternatively propose a more lightweight, finetuning-friendly method to integrate contextual information into SSD's framework, i.e., DiCSSD which uses multi-scale dilated convolution layers [31] shown as option 1 in Fig. 2. In DiCSSD, the context layers are fused via weighted sums (implemented as element-wise summations with learnable scaling parameters following a batch normalization layer) of multi-scale dilated convolution layers, where every original prediction layer is used individually, unlike DeCSSD that has explicit messaging between them.

DiCSSD rapidly expands the TRF sizes of each prediction layer, thus ensuring that every feature point within *sees* sufficiently large areas. If we choose a setting of 4 context layers in total for every prediction layer, then the TRF size in individual prediction layers will be $\times 2$, $\times 3$, $\times 4$ and $\times 5$ larger (Fig. 1). Intuitively, during feature map fusion the network collects a full set of visual cues that performs best in recognizing the object of interest. Note that the number of context layers is a hyper-parameter. In Sec. 6.1. we have set the value of that hyper-parameter to 4, based on cross-validation experiments.

## 4. Effective Receptive Fields

### 4.1. A Data-Driven Approach

We show how to obtain ERFs in Alg. 1 and Alg. 2 respectively. While inspired by [33], our algorithm differs from that [33] in that: 1) We evaluate our algorithms within an object detection framework, instead of a classification framework in [33]. Therefore, our algorithm selects ROIs with the highest activations over $K$ images for the given neuron, rather than top-$K$ images in [33]; 2) We visualize and validate ERF sizes of all conv layers in a VGG16 network (16 layers in total), which has been widely used in object detection methods, while [33] only provides ERF sizes of Places-CNN and ImageNet-CNN (5 layers in total), which have not been used much in object detection; 3) We further calculate ERF sizes of SSD additional conv layers using [20], based on the curve fitted from values of VGG16. Our result justifies why SSD has to extend its backbone from VGG16 with many conv layers and even more in its newest implementation [18]: If we take SSD300 as an example, the TRF size of *pool5* (VGG16's last layer) is 196 but its ERF is only about 104, which fills merely about 1/9 area of the original image. Therefore, the network is unable to learn fully the complex structures of very large objects.

Before running Alg. 1, we need to calculate the TRF size at the $i$-th layer, denoted as $H_t \times W_t$. After assigning $H^i = 1$ and $W^i = 1$ (according to the definition of RF), TRF size is calculated in reverse order of feature map size, i.e. recursively from $i$-th layer to the 1st layer:

$$H^{i-1} = (H^i - 1) \cdot s^i + l^i \cdot (k^i - 1) + 1$$
$$W^{i-1} = (W^i - 1) \cdot s^i + l^i \cdot (k^i - 1) + 1, \quad (1)$$

|  | conv2_1 | conv2_2 | pool2 | conv3_1 | conv3_2 | conv3_3 | pool3 | conv4_1 | conv4_2 | conv4_3 | pool4 | conv5_1 | conv5_2 | conv5_3 |
|---|---|---|---|---|---|---|---|---|---|---|---|---|---|---|
| TRF | 10 | 14 | 16 | 24 | 32 | 40 | 44 | 60 | 76 | 92 | 100 | 132 | 164 | 196 |
| ERF (Fitted [20]) | 5.1 | 9.9 | 12.0 | 19.5 | 25.8 | 31.4 | 34.0 | 43.3 | 51.4 | 58.6 | 62.1 | 74.5 | 85.4 | 95.4 |
| ERF (Data-Driven) | 6.9±0.1 | 9.8±0.2 | 11.1±0.2 | 16.4±0.3 | 21.4±0.4 | 25.4±0.6 | 27.2±0.8 | 36.8±0.8 | 45.7±1.0 | 54.1±1.6 | 56.9±1.8 | 77.6±2.9 | 94.4±4.4 | 104.2±7.8 |

Table 1: Comparison of sizes of ERFs and TRFs in a VGG16 network. Results of both a data-driven approach inspired by [33] and the fitted values based on theoretic $O(\sqrt{\text{TRF}})$ [20] are given.

**Algorithm 1** Discrepancy maps

**Input:** A pre-trained model $Z$, input image set $U \in \{I_n | n = 1, \ldots, N\}$, $I_n \in \mathcal{R}^{H_I \times W_I}$, occluder $o \in \mathcal{R}^{d \times d}$, stride $s$, layer $x$ with output size $H_f \times W_f \times l$
**Output:** Discrepancy maps $D \in \mathcal{R}^{H_I \times W_I \times l}$
1: Compute TRF size $H_t \times W_t$ at layer $x$ using Eqn. (1)
2: Locate only fully responsive ROI $Q \in \mathcal{R}^{H_Q \times W_Q}$ within area of rows $\in [\frac{H_t}{2}, H_I - \frac{H_t}{2}]$, cols $\in [\frac{W_t}{2}, W_I - \frac{W_t}{2}]$ /*To speed-up computation*/
3: Create empty discrepancy maps $D$
4: **while** $i \leq N$ **do**
5: In $Q \in U_i$, create occluded image set $V \in \{I_m | m = 1, \ldots, M\}$, $I_m \in \mathcal{R}^{H_I \times W_I}$ using $o$ with stride $s$
6: Record coordinates of each occluded area $P \in \{P_m | m = 1, \ldots, M\}$, $P_m \in \mathcal{R}^{d \times d}$
7: **while** $j \leq M$ **do**
8: Feedforward $U_i$ and $V_j$ through $Z$. At layer $x$, obtain activation maps respectively $A^x_{U_i}$ and $A^x_{V_j}$, both $\in \mathcal{R}^{H_f \times W_f \times l}$
9: **while** $k \leq l$ **do**
10: $D^k_{P_j} \leftarrow D^k_{P_j} + \sum |A^{x_k}_{U_i} - A^{x_k}_{V_j}|$
11: **end while**
12: **end while**
13: **end while**

**Algorithm 2** ERF

**Input:** Discrepancy maps $D \in \mathcal{R}^{H_I \times W_I \times l}$, TRF size $H_t \times W_t$ at layer $x$, threshold $\sigma$
**Output:** ERF size $H_e \times W_e$
1: **while** $i \leq l$ **do**
2: $P_i \leftarrow \underset{(m,n)}{\arg\max}\, D^i$ /*Calibration*/
3: $E_i \leftarrow$ ROI centered at $P_i$ in $D^i$, $E_i \in \mathcal{R}^{H_f \times W_f}$
4: $S_i \leftarrow \sqrt{\sum [E_i \geq \sigma * \mu(\text{vec}(E_i))]}$ /*[...] are the Iverson brackets*/
5: **end while**
6: $H_e \leftarrow \mu(S)$
7: $W_e \leftarrow \mu(S)$

where $s^i$, $l^i$, $k^i$ denote convolution stride, dilation stride and kernel size at the $i$-th layer respectively. Finally, we have $H_t = H^1$, $W_t = W^1$.

The ERF sizes of VGG16 are shown in Table 1. We compute ERF (Alg. 2) with a total of $K = 300$ images and threshold $\sigma = 1.0$. With $K = 100$, the computed ERF values across all layers show at most ±0.1 difference from the corresponding values computed with $K = 300$, therefore it is safe to use only 100 images.

### 4.2. Analysis and Visualization

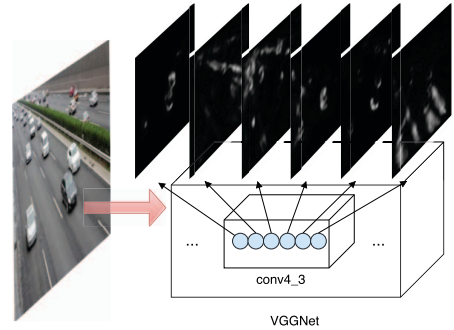

Figure 4: Example discrepancy maps of *conv4_3*. Each neuron responses differently with semantics from input image. The raster effects on discrepancy maps are due to *conv4_3*'s limited ERF size around 54.

We show some example discrepancy maps of *conv4_3* in Fig. 4. Each discrepancy map shows regions to which certain neuron are most responsive. Those regions are semantic (car roofs/shields, highway shrubs, lanes, etc.) and thus can offer us more insights into how neurons respond to our input image (similar to *emergence of detectors* in [33]).

After obtaining $K$ discrepancy maps, for all neurons in a certain layer we extract ROIs centered at the points with max activation (named *Calibration* in Alg. 2). Subsequently, ROIs are averaged over all neurons, after which we will see a typical 2D-Gaussian shape in accordance with [20], as shown in Fig. 3.

Fig. 5 demonstrates our findings about ERF: In Fig. 5(a), $\frac{\text{ERF}}{\text{TRF}}$ becomes smaller at top layers which is again in accordance to [20]. In addition, we fit a linear curve of ERF vs. $\sqrt{\text{TRF}}$ sizes in Fig. 5(b), where the variations of two lines are mainly due to the pooling layers interleaved before *conv5_3* in VGG16 network, as well as the data noise introduced when calculating ERFs. Last but not least, we compare ERF sizes against grid scales of SSD at different prediction layers in Fig. 5(c), from which we realize that the sizes of SSD's default boxes are not sufficiently large to include context, especially for small objects (*conv4_3*).

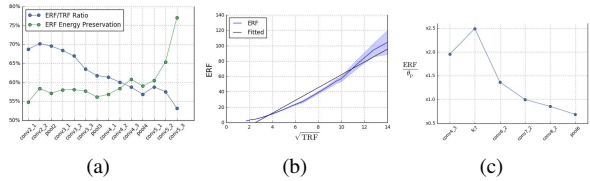

(a) (b) (c)

Figure 5: (a) As the network goes deeper, ERF retains more intensity, and surprisingly has lower ratio against its TRF counterpart. (b) A likely-linear relationship between ERF and $\sqrt{\text{TRF}}$ sizes. (c) Fitted ERF sizes to corresponding grid scales $\theta_p$ at different prediction layers.

## 5. Training

We follow the same training process as [17]: After generating default boxes, we match each of the ground truth boxes to the best overlapped default boxes with Jaccard overlap higher than 0.5. For the rest of default boxes, which have been left unmatched, we select a subset of them based on confidence loss while keeping the ratio of matched to unmatched boxes to 1:3, which keeps a balance between number of positive and negative proposals. Afterwards, our objective function minimizes regression loss using Smooth $\ell_1$ and classification loss using Softmax. Note that SSD was updated with a new expansion data augmentation trick that has been shown to boost the performance a lot in small object detection [18]. In this paper, we follow the newest data augmentation expansion trick in our experiments on VOC, while keep using the original data augmentation technique [17] on DETRAC.

Both implementations [17, 18] use the entire original image as input, and then randomly sample patches that have the *minimum* Jaccard overlap with one of the ground truth objects. Multiple minimum Jaccard overlap thresholds have been used, including 0.1, 0.3, 0.5, 0.7 and 0.9, each with 50 maximum trials. After doing so, the statistical distribution of training sample scales fed into the network is expected to be more *equalized*.

## 6. Experiments

### 6.1. Ablation Study

In Fig. 6(a), we study how mAP changes with different number of context layers for DiCSSD, as measured on the DETRAC dataset. Note that the overall mAP score is not necessarily indicative of the mAP value for small object detection. We note that with higher number of context layers, the overall mAP score of DiCSSD drops dramatically while memory requirements increase. We set the number of context layers equal to 4, based on cross-validation on the

DETRAC dataset, and we used the same number (4 context layers) for the VOC 2007 dataset. In addition, we conducted experiments under different settings of batch normalization, scaling, and context layer fusion method and show the results in Fig. 6(b). The *golden* model compared in Fig. 6(b) has been trained with a full combination of batch normalization + scaling + sum.

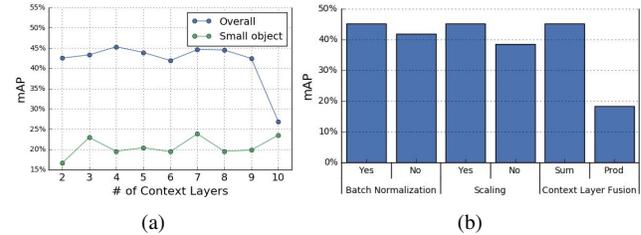

(a) (b)

Figure 6: We perform our ablation study on DETRAC (Sec. 6.2) to test, for our DiCSSD method: (a) mAP under different number of context layers. (b) Effects of different settings with batch normalization, scaling and context layers fusion.

### 6.2. Our Results

Similar to SSD, we load a pre-trained model of VGG16 on the ILSVRC CLS-LOC dataset [26] and use the átrous algorithm [12] to convert *fc6* and *fc7* into conv layers. In addition, we use SGD with initial learning rate $10^{-3}$, 0.9 momentum, 0.0005 weight decay and 32 batch size. The learning rate decay policy, however, is different depending on the dataset used in experiments: for DETRAC, we use an initial learning rate $10^{-3}$ at the first 40k iterations, then continue training with learning rate $10^{-5}$ until the model is fully converged at iteration 60k. For PASCAL VOC 2007, we use the same training policy with the newest implementation of SSD [18], which uses initial learning rate $10^{-3}$ at first 80k iterations and continues training two rounds of 20k iterations with learning rate $10^{-4}$, $10^{-5}$ respectively. For MS-COCO, we train our model with learning rate $10^{-3}$ for the first 160k iterations, followed by two rounds of 40k iterations with learning rate $10^{-4}$ and $10^{-5}$ respectively.

**DETRAC.** DETRAC is a challenging real-world vehicle detection dataset [30], with a total of 10 hours of videos. It consists of 60 sequences in *train* set and 40 in *test* set, which include significant differences in vehicle categories, weather, scales, occlusion ratios, and truncation ratios.

We note that ground truth for the DETRAC *test* set has not been released yet. Since we needed ground truth for our quantitative experiments, we opted to split the original 60 sequences of the DETRAC *train* set into 48 sequences that we used for training, and 12 sequences that we used as our test set. Our test set consisted of the 12 sequences numbered by *20034, 20063, 39851, 40131, 40191, 40243, 40871, 40962, 40992, 41063, 63521, 63562*. Moreover, we subsampled frames with step size 10 in all sequences, leading to a total of 6349 training images and 1880 test images. To have a comprehensive evaluation of the proposed networks, we also assign to each ground truth bounding box new annotations for scale, occlusion and three difficulty levels, as shown in Table 4.

| Method | Network | Overall | Category | | | | Difficulty | | | Occlusion | | | Scale | | | Weather | | | |
|---|---|---|---|---|---|---|---|---|---|---|---|---|---|---|---|---|---|---|---|
| | | | Car | Van | Bus | Others | Easy | Normal | Hard | No | Partial | Heavy | Small | Medium | Large | Sunny | Rainy | Night | Cloudy |
| DiCSSD533 | VGG16 | **55.6** | 63.2 | **52.1** | **81.2** | **46.3** | **63.2** | **41.0** | **12.2** | 57.5 | **47.4** | **26.7** | 35.6 | **79.9** | **84.4** | 61.8 | **70.6** | **53.9** | 69.8 |
| DeCSSD533 | VGG16 | 51.5 | **64.3** | 47.9 | 71.9 | 14.3 | 60.6 | 40.9 | 10.2 | **59.2** | 44.2 | 24.3 | **39.2** | 76.5 | 68.5 | **64.0** | 64.8 | 52.8 | **72.2** |
| SSD533 [17] | VGG16 | 50.3 | 58.7 | 45.6 | 75.8 | 37.0 | 61.3 | 34.1 | 7.1 | 56.5 | 38.9 | 17.4 | 31.1 | 74.2 | 81.7 | 55.6 | 64.0 | 52.5 | 63.8 |
| DiCSSD300 | VGG16 | **45.3** | **53.0** | **39.5** | **71.5** | 19.4 | **61.9** | **25.0** | **9.8** | **50.9** | 32.5 | **24.0** | **19.6** | **72.5** | **78.7** | **48.5** | **60.7** | **48.1** | **54.6** |
| DeCSSD300 | VGG16 | 40.2 | 47.8 | 37.4 | 64.6 | 11.4 | 57.3 | 21.1 | 7.2 | 47.8 | 30.3 | 17.6 | 14.7 | 67.8 | 63.3 | 41.2 | 55.1 | 45.7 | 53.0 |
| SSD300 [17] | VGG16 | 35.1 | 43.5 | 24.8 | 50.2 | 4.5 | 53.2 | 19.0 | 6.5 | 42.8 | 26.6 | 16.2 | 14.2 | 60.2 | 59.4 | 42.0 | 42.3 | 39.4 | 51.6 |
| SSD300 [17] | ResNet152 | 35.2 | 42.3 | 33.5 | 54.3 | 4.5 | 52.8 | 17.5 | 7.1 | 41.1 | **34.7** | 17.6 | 6.8 | 62.9 | 60.2 | 36.8 | 50.8 | 30.5 | 44.4 |
| SSD300 [17] | ResNet101 | 39.1 | 42.8 | 32.6 | 69.5 | **30.2** | 55.4 | 16.4 | 7.1 | 43.2 | 30.4 | 17.5 | 8.1 | 63.1 | 69.5 | 40.0 | 48.0 | 44.2 | 46.5 |
| SSD300 [17] | ResNet50 | 25.1 | 42.4 | 30.5 | 59.8 | 26.9 | 46.1 | 4.8 | 0.0 | 35.9 | 0.4 | 0.0 | 7.4 | 63.2 | 63.8 | 44.7 | 0.0 | 0.0 | 0.0 |

Table 2: Evaluation of the proposed networks on DETRAC dataset.

| Method | Network | mAP | aero | bike | bird | boat | bottle | bus | car | cat | chair | cow | table | dog | horse | mbike | person | plant | sheep | sofa | train | tv |
|---|---|---|---|---|---|---|---|---|---|---|---|---|---|---|---|---|---|---|---|---|---|
| DiCSSD300* | VGG16 | **78.1** | **82.2** | **85.4** | 76.5 | 69.8 | **51.1** | 86.4 | **86.4** | 88.0 | 61.6 | 82.7 | 76.4 | **86.5** | **87.9** | **85.7** | 78.8 | **54.2** | 76.9 | 77.6 | **88.9** | **78.2** |
| DeCSSD300* | VGG16 | 77.6 | 79.9 | 84.7 | 76.4 | **70.2** | 48.2 | 86.5 | 86.1 | **88.9** | **61.7** | **83.1** | 76.8 | 86.1 | 87.4 | 85.3 | 78.8 | 52.0 | 77.0 | 79.1 | 87.0 | 77.2 |
| SSD300* [18] | VGG16 | 77.5 | 79.5 | 83.9 | 76.0 | 69.6 | 50.5 | **87.0** | 85.7 | 88.1 | 60.3 | 81.5 | **77.0** | 86.1 | 87.5 | 84.0 | **79.4** | 52.3 | **77.9** | **79.5** | 87.6 | 76.8 |
| DSSD321 [6] | ResNet101 | 78.6 | 81.9 | 84.9 | 80.5 | 68.4 | 53.9 | 85.6 | 86.2 | 88.9 | 61.1 | 83.5 | 78.7 | 86.7 | 88.7 | 86.7 | 79.7 | 51.7 | 78.0 | 80.9 | 87.2 | 79.4 |
| ION [1] | VGG16 | 75.6 | 79.2 | 83.1 | 77.6 | 65.6 | 54.9 | 85.4 | 85.1 | 87.0 | 54.4 | 80.6 | 73.8 | 85.3 | 82.2 | 82.2 | 74.4 | 47.1 | 75.8 | 72.7 | 84.2 | 80.4 |
| R-FCN [14] | ResNet101 | 80.5 | 79.9 | 87.2 | 81.5 | 72.0 | 69.8 | 86.8 | 88.5 | 89.8 | 67.0 | 88.1 | 74.5 | 89.8 | 90.6 | 79.9 | 81.2 | 53.7 | 81.8 | 81.5 | 85.9 | 79.9 |
| Faster [11] | ResNet101 | 76.4 | 79.8 | 80.7 | 76.2 | 68.3 | 55.9 | 85.1 | 85.3 | 89.8 | 56.7 | 87.8 | 69.4 | 88.3 | 88.9 | 80.9 | 78.4 | 41.7 | 78.6 | 79.8 | 85.3 | 72.0 |
| Fast [8] | VGG16 | 70.0 | 77.0 | 78.1 | 69.3 | 59.4 | 38.3 | 81.6 | 78.6 | 86.7 | 42.8 | 78.8 | 68.9 | 84.7 | 82.0 | 76.6 | 69.9 | 31.8 | 70.1 | 74.8 | 80.4 | 70.4 |
| Faster [25] | VGG16 | 73.2 | 76.5 | 79.0 | 70.9 | 65.5 | 52.1 | 83.1 | 84.7 | 86.4 | 52.0 | 81.9 | 65.7 | 84.8 | 84.6 | 77.5 | 76.7 | 38.8 | 73.6 | 73.9 | 83.0 | 72.6 |

Table 3: Detection results on PASCAL VOC2007 *test* set. All models were trained with VOC2007 *trainval* set + VOC2012 *trainval* set. (C)SSD300* and SSD512* are the latest SSD models with the new expansion data augmentation trick [18].

We compare the performance of CSSD with SSD in Table 2. In that table, (C)SSD533 indicate models trained with input size $533 \times 300$, which keeps the same aspect ratio with the original input size $940 \times 540$, and has been found to greatly boost the mAP of (C)SSD300 (50.3% vs. 35.1% for SSD300). Our results show that both DiCSSD and DeCSSD outperform SSD with two different input resolutions, while DiCSSD significantly improves SSD with higher mAP than DeCSSD ($+10.2\%$ vs. $+5.1\%$ to SSD300, $+5.3\%$ vs. $+1.2\%$ to SSD533). In small object detection, DiCSSD300 has been found particularly effective, with mAP increase $> 5\%$ over both DeSSD300 and SSD300. Note that, although DeCSSD533 achieves the highest mAP in small object detection among the three, its performance on large object is down to the lowest 68.5%. This happens because, given high resolution images, top prediction layers may contain much richer feature maps, but directly reusing and fusing those maps into a single map may overweigh the training loss at the first prediction layer. Consequently, while DeCSSD outperforms the other methods in small object detection, its discriminative capabilities for large objects have been greatly constrained.

| Scale | Small<br>0-50 pixels | Medium<br>50-150 pixels | Large<br>> 150 pixels |
|---|---|---|---|
| Occlusion | No<br>< 1% | Partial<br>1% − 50% | Heavy<br>> 50% |
| Difficulty | Easy<br>Medium or large object,<br>no occlusion,<br>$0\% \leq$ truncation $< 15\%$ | Normal<br>Partial occlusion, or<br>$15\% \leq$ truncation $< 30\%$ | Hard<br>Heavy occlusion, or<br>truncation $\geq 50\%$. |

Table 4: New annotations created for DETRAC, where each of them has been evaluated in Table 2. Bounding boxes with undefined conditions in difficulty are defaulted to *Normal*.

| Method | mAP | Network | FPS | | | Input Resolution |
|---|---|---|---|---|---|---|
| | | | K40 | GTX 980 | Titan X (Pascal) | |
| SSD300* [18] | 77.5 | VGG16 | 16.2 | 36.2 | 46 | 300x300 |
| DiSSD300* | 78.1 | VGG16 | 12.2 | 24.3 | 40.8 | 300x300 |
| DeSSD300* | 77.6 | VGG16 | 8.7 | 18.4 | 39.8 | 300x300 |
| DSSD [6] | 78.6 | ResNet101 | - | - | 9.5 | 321x321 |
| YOLO [23] | 66.4 | VGG16 | - | - | 21 | 448x448 |
| Faster R-CNN [8] | 73.2 | VGG16 | - | - | 7 | ∼1000x600 |
| Faster R-CNN [11] | 76.4 | ResNet101 | - | - | 2.4 | ∼1000x600 |
| R-FCN [14] | 80.5 | ResNet101 | - | - | 9 | ∼1000x600 |

Table 5: Speed and accuracy comparisons on VOC2007.

In summary, DiCSSD was found to be the most effective compared to others in DETRAC, with both low and high-resolution input size. Later on, we will see that the run-time speed of DiCSSD is comparable to SSD, which proves the efficiency of our proposed method.

**PASCAL VOC 2007.** In this dataset, we train each model with a union of VOC2007 *trainval* set and VOC 2012 *trainval* sets, and evaluate on VOC 2007 *test* set. Again, our results in Table 3 indicate that both DiCSSD and DeCSSD outperform SSD ($+0.6\%$ vs. $+0.1\%$). We note that the performance of SSD has been improved greatly with its new expansion data augmentation trick [18]. Still, our proposed context layers, applied on top of this improved SSD, further boost performance and take little memory consumption. It is noteworthy that among all the 20 categories evaluated, mAPs of DiCSSD are higher than those of SSD in 15 categories.

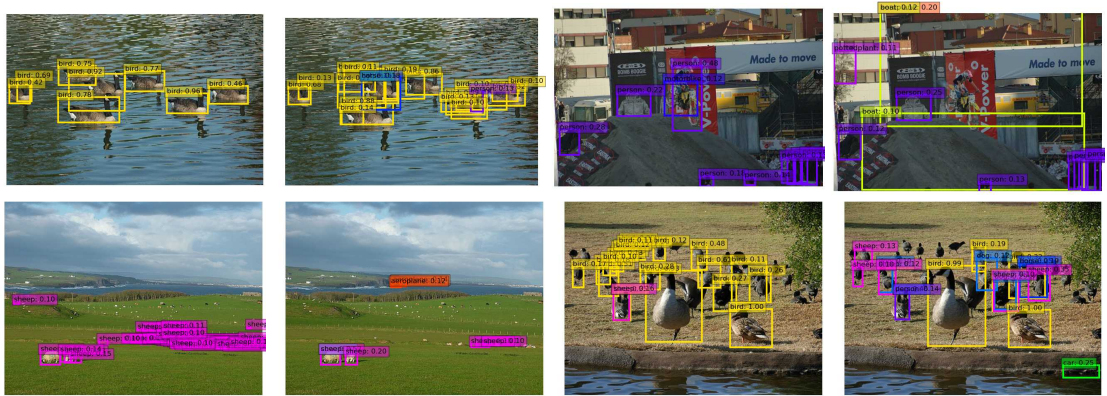

Figure 7: Curated examples of DiCSSD (left) and SSD (right) on VOC07 *test* set.

| Method | Train Set | Network | FPS | Avg. Precision, IoU: | | | Avg. Precision, Area: | | | Avg. Recall, #Dets: | | | Avg. Recall, Area: | | |
|---|---|---|---|---|---|---|---|---|---|---|---|---|---|---|---|
| | | | | 0.5:0.95 | 0.5 | 0.75 | S | M | L | 1 | 10 | 100 | S | M | L |
| DiCSSD300* | trainval35k | VGG16 | 40.8 | **26.9** | **46.3** | **27.7** | **8.2** | **27.5** | **43.4** | **25.0** | **37.3** | **39.8** | **15.4** | **43.1** | **60.0** |
| SSD300* [18] | trainval35k | VGG16 | 46 | 25.1 | 43.1 | 25.8 | 6.6 | 25.9 | 41.4 | 23.7 | 35.1 | 37.2 | 11.2 | 40.4 | 58.4 |
| SSD300 [17] | trainval35k | VGG16 | 46 | 23.2 | 41.2 | 23.4 | 5.3 | 23.2 | 39.6 | 22.5 | 33.2 | 35.3 | 9.6 | 37.6 | 56.5 |
| DSSD321 [6] | trainval35k | ResNet101 | 9.5 | 28.0 | 46.1 | 29.2 | 7.4 | 28.1 | 47.6 | 25.5 | 37.1 | 39.4 | 12.7 | 42.0 | 62.6 |
| R-FCN [14] | trainval | ResNet101 | 9 | 29.9 | 51.9 | - | 10.8 | 32.8 | 45.0 | - | - | - | - | - | - |
| ION [1] | train | VGG16 | - | 23.6 | 43.2 | 23.6 | 6.4 | 24.1 | 38.3 | 23.2 | 32.7 | 33.5 | 10.1 | 37.7 | 53.6 |
| Faster [25] | trainval | VGG16 | 7 | 21.9 | 42.7 | - | - | - | - | - | - | - | - | - | - |
| Fast [8] | train | VGG16 | - | 19.7 | 35.9 | - | - | - | - | - | - | - | - | - | - |

Table 6: MS-COCO *test-dev2015* detection results.

**MS-COCO 2015.** In order to evaluate CSSD on a more general, large-scale object detection dataset, we compare our proposed model with both SSD , DSSD and many others on MS-COCO 2015 dataset. To directly compare CSSD with SSD, we use the same *trainval35k* training set [1] and follow the same training policy to SSD300* [18]. In Table 6, we show our detection results on MS-COCO [16] *test-dev2015* set. Both DiSSD300* and SSD300* use the new expansion data augmentation trick [18], while SSD300 is with the original SSD implementation [17].

Our proposed network DiCSSD has been shown to greatly boost the performance of SSD (+3.2% given IoU=0.5), and even better than DSSD (+0.1% given IoU=0.5) which uses ResNet101 as its base network but only achieves 9.5fps (compared to 40.8 fps of ours). We also note that for small object, DiCSSD achieves better precision than the newest SSD (8.2% vs. 6.6%), in addition to the highest recall (15.4%) among all the benchmarked methods.

**Inference time.** In Table 5, we compare the speed and accuracy of benchmarked models on PASCAL VOC 2007. The proposed DiCSSD network achieves the highest mAP, while maintaining a real-time speed of 40.8 fps. Therefore, DiCSSD has been the most effective model among all models compared. Note that we calculate fps with batch size = 1, preprocessing time counted and cuDNN [2] enabled (v5.1 for Titan X, v4 for K40 and GTX 980).

# 7. Conclusion

We have presented a new deep learning based object detection algorithm, called context-aware single-shot detector (CSSD). CSSD incorporates object context modeling by integrating dilated convolution layers or deconvolution layers into the state-of-the-art SSD algorithm. Our experiments show significant improvement in object detection accuracy compared to SSD on the MS-COCO and DETRAC dataset, where the improvement is much more pronounced for small scale objects. Algorithms for obtaining effective receptive fields (ERFs) are also presented, with the empirical study results further supporting our hypothesis that the context layers are important for achieving superior detection accuracy, especially for small objects.

# Acknowledgement

This work was partially supported by National Science Foundation grants IIS 1565328 and IIP 1719031.